\documentclass[sigconf,screen]{acmart}

\usepackage{algorithm}
\usepackage{algorithmic}
\usepackage{booktabs} 
\usepackage{colortbl} 
\usepackage{enumitem}
\usepackage{makecell}
\usepackage{multirow} 
\usepackage{multicol}
\usepackage{xcolor}  

\copyrightyear{2026}
\acmYear{2026}
\setcopyright{cc}
\setcctype{by}
\acmConference[MM '26]{Proceedings of the 34th ACM International Conference on Multimedia}{November 10--14, 2026}{Rio de Janeiro, Brazil}
\acmBooktitle{Proceedings of the 34th ACM International Conference on Multimedia (MM '26), November 10--14, 2026, Rio de Janeiro, Brazil}
\acmDOI{10.1145/3767308.3835789}
\acmISBN{979-8-4007-2213-4/2026/11}

\begin{document}

\title{\textbf{MoECa: Aligning Feature Reuse with Expert Decomposition in Diffusion Transformers}}

\author{Maoliang Li}
\email{maoliang.li@stu.pku.edu.cn}
\authornote{Both authors contributed equally to this research.}
\affiliation{%
  \institution{Peking University}
  \city{Beijing}
  \country{China}}

\author{Haojing Chen}
\email{2023090913005@std.uestc.edu.cn}
\authornotemark[1]
\affiliation{%
  \institution{University of Electronic Science and Technology of China}
  \city{Chengdu}
  \country{China}}

\author{Jiayu Chen}
\email{jiayu.chen.25@stu.pku.edu.cn}
\affiliation{%
  \institution{Peking University}
  \city{Beijing}
  \country{China}}

\author{Zihao Zheng}
\email{zhengzihao@stu.pku.edu.cn}
\affiliation{%
  \institution{Peking University}
  \city{Beijing}
  \country{China}}

\author{Xinhao Sun}
\email{sunxinhao5513@gmail.com}
\affiliation{%
  \institution{Peking University}
  \city{Beijing}
  \country{China}}

\author{Hailong Zou}
\email{zouhailong26@stu.pku.edu.cn}
\affiliation{%
  \institution{Peking University}
  \city{Beijing}
  \country{China}}

\author{Xiang Chen}
\authornote{Corresponding author.}
\email{xiang.chen@pku.edu.cn}
\affiliation{%
  \institution{Peking University}
  \city{Beijing}
  \country{China}}

\begin{abstract}
Diffusion Transformers with Mixture-of-Experts (DiT-MoE) improve model capacity under sparse activation, but diffusion inference is still bottlenecked by redundant computation across timesteps. Existing caching methods mainly operate at the token level, which becomes suboptimal in DiT-MoE because each token update is internally decomposed into multiple routed expert branches. Our analysis shows that cross-timestep redundancy in DiT-MoE is better characterized at the expert-branch level than at the whole-token level. Based on this observation, we propose MoECa, a fine-grained caching framework that performs branch-level feature reuse across timesteps. MoECa further introduces expert-aware adaptive control and synchronized cache updates across MoE and attention paths to maintain stable intermediate states. Experiments on multiple DiT-MoE models show a favorable speed--quality trade-off, with up to 2.93$\times$ speedups while preserving generation quality.
\end{abstract}



\maketitle



\section{Introduction}

\begin{figure}[t!]
    \centering
    \includegraphics[width=3in]{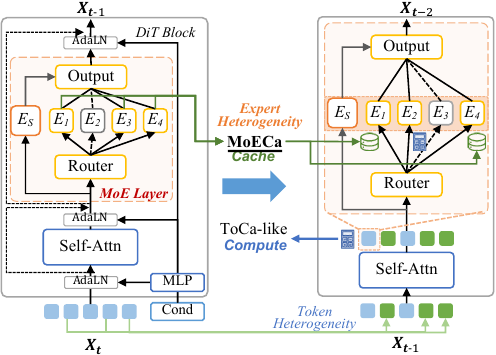}
    \vspace{-4mm}
    \caption{Conceptualization of: (a) expert heterogeneity in DiT-MoE; and (2) misalignment between token-level cache and MoE models.}
    \label{fig:1}
    \vspace{-4mm}
\end{figure}

Diffusion models, especially Diffusion Transformers (DiTs)~\cite{peebles2023DiT}, have become a dominant paradigm for visual generation due to their strong scalability and generation quality.
    As model sizes continue to grow, researchers~\cite{shi2025diffmoe,fei2024DiTMoE,cai2025hidreami1,shi2026mamoda,akiti2026nucleusimage} have begun to introduce sparse Mixture-of-Experts (MoE)~\cite{shazeer2017sparselygated} into DiT, as illustrated in Fig.~\ref{fig:1}, to improve model capacity while keeping the number of active parameters manageable.
    More fundamentally, the effect of MoE is not limited to the capacity gain from sparse activation.
    By replacing the original dense transformation with multiple expert branches, MoE reorganizes the feature evolution process of each token into the joint progression of multiple expert branch features.

Beyond model capacity, the inference efficiency of diffusion models is still fundamentally constrained by the iterative denoising process. 
    In addition to reducing the number of sampling steps through improved samplers~\cite{song2020denoising} or distillation~\cite{yin2024one}, feature caching has become one of the most widely adopted acceleration paradigms for diffusion inference.
    The underlying motivation is that features at adjacent timesteps often exhibit considerable similarity, making it possible to cache and reuse previously computed features to avoid redundant computation. 
    Existing studies further show that temporal variation is heterogeneous across different granularities, such as layers~\cite{liu2024timestep}, blocks~\cite{zhang2025blockdance}, and tokens~\cite{zou2024accelerating,selvaraju2024fora}, and that selective reuse strategies can therefore achieve favorable speed-quality trade-offs.
    In dense DiTs, token-level reuse is a natural design choice because token heterogeneity is one of the primary sources of temporal redundancy in visual representations.
    However, in MoE models, token features are further decomposed into multiple expert branches.
    With token as the atomic unit for selective update, the branch-level heterogeneity introduced by expert decomposition is ignored.

\begin{figure}[t!]
    \centering
    \includegraphics[width=3in]{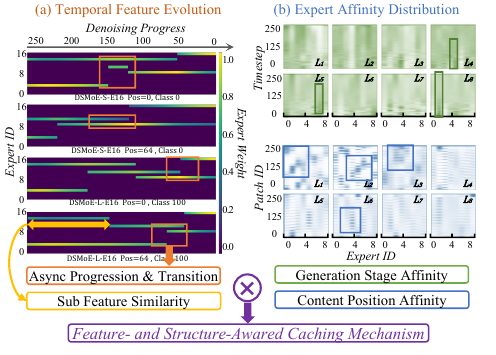}
    \vspace{-4mm}
    \caption{Preliminary Observation: (a) Temporal progression of MoE feature for example tokens; (b) Expert activation across Image Patch and Timestep.}
    \vspace{-4mm}
    \label{fig:2}
\end{figure}

To this end, the temporal evolution of noisy features over experts and expert branches becomes the key to understanding why token-level reuse is suboptimal in DiT-MoE. 
    We first present two preliminary observations in Fig.~\ref{fig:2}.
    From the perspective of cross-timestep feature evolution, Fig.~\ref{fig:2}(a) shows that the branch-wise feature evolution within the same token becomes highly heterogeneous, while each individual branch still exhibits noticeable temporal similarity. 
    From the perspective of expert specialization, Fig.~\ref{fig:2}(b)\footnote{Illustration style and sampling scripts are adapted from DiT-MoE\cite{fei2024DiTMoE}} shows that experts exhibit various response affinities over content positions and denoising stages.
    Consequently, the information carried by different experts is not identical, which further suggests that different experts should not be treated as equally sensitive to caching error.
    Taken together, the internal representation evolution in DiT-MoE simultaneously exhibits \emph{cross-timestep similarity} and \emph{cross-expert / cross-branch heterogeneity}.
    Therefore, as illustrated in Fig.~\ref{fig:1}, existing token-level caching leads to suboptimal reuse decisions in DiT-MoE.
    On one hand, recomputing the entire token to avoid error accumulation forces stable branches to be unnecessarily recomputed, introducing redundancy.
    On the other, reusing the entire token for more acceleration causes rapidly changing branches to be excessively reused, resulting in larger error.

To address this problem, we propose \textbf{MoECa}, a feature- and structure-aware caching mechanism for MoE-based DiTs. 
    Instead of treating each token as the atomic caching unit, MoECa pushes the caching granularity down to the expert-branch level, so that caching decisions are aligned with the internal computation decomposition structure of MoE and its feature evolution pattern. 
    On top of this design, MoECa builds a branch-wise caching and update framework, introduces an expert-aware adaptive control mechanism, and further coordinates the MoE path and the attention path to reduce inference overhead while preserving generation quality as much as possible.
Our contributions are summarized as follows:
    \textbf{(1)} We analyze the cross-timestep evolution of features in DiT-MoE from the perspective of expert decomposition and specialization, and reveal the structural mismatch between caching granularity and feature representations of MoE.
    \textbf{(2)} We propose \textbf{MoECa}, a fine-grained caching framework for DiT-MoE, which enables more efficient feature reuse through expert-branch-level caching and update.
    \textbf{(3)} We evaluate MoECa on multiple popular DiT-MoE models, demonstrating a favorable speed--quality trade-off across model families and generation settings.

\section{Backgrounds}

\subsection{Diffusion Models and Transformers}

\textbf{Diffusion models} are built upon two complementary stochastic processes: a forward noising process and a reverse denoising process. 
    In the forward process, Gaussian perturbations are progressively injected into a clean sample until it approaches an isotropic Gaussian distribution. 
    In the reverse process, a learnable network iteratively removes noise and maps a noisy sample back to the data manifold. 
    Let $t$ denote the diffusion timestep and $\beta_t$ the variance schedule.
    The reverse transition can be written as
\begin{equation}
\label{eq:reverse_process_dit_section}
    p_\theta(x_{t-1}\mid x_t)=
    \mathcal{N}(x_{t-1};\frac{1}{\sqrt{\alpha_t}}(x_t-\frac{1-\alpha_t}{\sqrt{1-\bar{\alpha}_t}}\,\epsilon_\theta(x_t,t)),\beta_t\mathbf{I}),
\end{equation}
    where $\alpha_t=1-\beta_t$ and $\bar{\alpha}_t=\prod_{i=1}^{t}\alpha_i$. Here, $\epsilon_\theta(\cdot)$ is the denoiser parameterized by $\theta$, which predicts the noise component from $(x_t,t)$. The denoiser is repeatedly invoked over $T$ timesteps during sampling, dominating both generation quality and inference cost.

Recent diffusion systems commonly instantiate $\epsilon_\theta$ with a Transformer, yielding \textbf{Diffusion Transformers} (DiT). 
    A DiT stacks $L$ blocks, each typically containing self-attention and MLP sublayers.
\begin{equation}
\label{eq:dit_compose}
\mathcal{G}=g_1\circ g_2\circ\cdots\circ g_L,\quad
g_l=\mathcal{F}_{\mathrm{SA}}^{l}\circ\mathcal{F}_{\mathrm{MLP}}^{l}.
\end{equation}
    The input at timestep $t$ is represented as a token sequence $\mathbf{x}_t=\{x_i\}_{i=1}^{H\times W}$, where each token corresponds to an image (or latent) patch. For a generic sublayer $f$, the block-level update is usually implemented in residual form, $\mathcal{F}(\mathbf{x})=\mathbf{x}+\mathrm{AdaLN}\circ f(\mathbf{x})$,
    where AdaLN modulates normalized activations according to diffusion timestep (and optional condition signals), enabling the model to adapt computation across denoising stages. 

\subsection{Feature Caching for Diffusion Acceleration}

Feature caching accelerates DiT models by reusing intermediate features across nearby timesteps.
    Let $\{t,t+1,\dots,t+N-1\}$ be a window of $N$ adjacent denoising steps. 
    At the first step $t$ (refresh step), the model executes normally and writes intermediate features into cache memory. 
    Denoting the cache at layer index $l$ as $\mathcal{C}[l]$, cache is assigned as:
\begin{equation}
\label{eq:dit_cache_refresh}
    \mathcal{C}[l] := 
        \mathcal{F}_l\!\left(\mathbf{x}_t^{l-1}\right), \qquad l=1,\dots,L,
\end{equation}
    For each following step $t+i$ with $i\in\{1,\dots,N-1\}$, naive caching directly reuses the stored intermediate feature:
\begin{equation}
\label{eq:dit_cache_reuse}
    \mathcal{F}_l\!\left(\mathbf{x}_{t+i}^{l-1}\right) :=
        \mathcal{C}[l], \qquad l=1,\dots,L.
\end{equation}
    In practice, one refresh step is amortized over several reuse steps to reduce attention/MLP computation, but overly long reuse windows can cause cache staleness and quality degradation.

From the perspective of feature granularity, caching methods can be divided into step-, block-, layer-, and token-level, by hierarchy of re-use decision in DiT architecture.
    Finer reuse better preserves quality but requires higher control overhead. 
    ToCa is a representative token-level method that selectively recomputes only high-variation tokens, which effectively exploits feature heterogeneity across various tokens.

\begin{figure}[t!]
    \centering
    \includegraphics[width=3in]{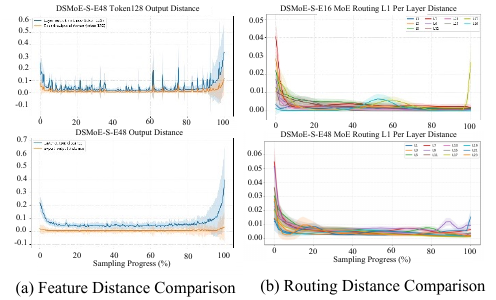}
    \vspace{-4mm}
    \caption{Temporal Feature Evolution Analysis: (a) Distance between adjacent timesteps of token- and branch-level features. The upper figure sets distance to 0 for branch transition. (b) Distance between adjacent timesteps of routing output.}
    \vspace{-4mm}
    \label{fig:3}
\end{figure}

\subsection{MoE and DiT-MoE}
Mixture-of-Experts (MoE) replaces the dense FFN in Transformer structure with multiple routed experts and a router. 
    For token input $u_t$, the standard MoE FFN output (without shared experts) is
\begin{equation}
\label{eq:moe_output_base}
    h_t = u_t + 
        \sum\nolimits_{i=1}^{N_r}g_{i,t}\,\mathrm{FFN}^{(r)}_i(u_t),
\end{equation}
    where $N_r$ is the number of routed experts.
Since such structure is inherently compatible with the DiT paradigm, DSMoE~\cite{liu2025efficienttrainingdiffusionmixtureofexperts}, a practical DiT-MoE training recipe with sparse routed and shared experts, and DiT-MoE~\cite{fei2024DiTMoE} introduce MoE to DiT models without significant modification to model structure.
Following DeepSeek MoE~\cite{dai2024deepseekmoe}, shared experts are further introduced in addition to routed experts, so the output of original FFN becomes:
\begin{equation}
\label{eq:moe_output}
    h_t = u_t
        +\sum\nolimits_{i=1}^{N_s}\mathrm{FFN}^{(s)}_i(u_t)
        +\sum\nolimits_{i=1}^{N_r}g_{i,t}\,\mathrm{FFN}^{(r)}_i(u_t),
\end{equation}
    where $N_s$ and $N_r$ are the numbers of shared experts and routed experts, $\mathrm{FFN}^{(s)}_i(\cdot)$ denotes the $i$-th shared expert, and $\mathrm{FFN}^{(r)}_i(\cdot)$ denotes the $i$-th routed expert.

Overall, the role of MoE in DiT is not limited to sparse activation for a better capacity--efficiency trade-off. 
    More importantly, token-wise routing decomposes the original dense FFN update into multiple expert-conditioned branch updates, so the representation evolution of each token is no longer governed by a single homogeneous computation path. 
    Since experts in DiT-MoE often exhibit different preferences over spatial positions and denoising stages, MoE also changes how noisy features are internally updated and evolve across timesteps. 
    As a result, this structural change is directly relevant to feature reuse during diffusion inference, where whole-token reuse decisions may no longer align well with the branch-wise computation induced by expert decomposition.

\section{Analysis}
To understand cross-timestep feature reuse pattern in DiT-MoE, we analyze how noisy features evolve across timesteps under MoE decomposition, and discuss its misalignment with token-level caching.

\begin{figure}[t!]
    \centering
    \includegraphics[width=3in]{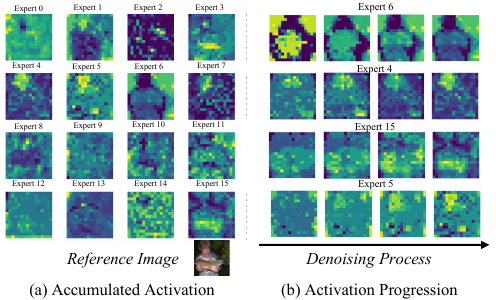}
    \vspace{-4mm}
    \caption{Expert Activation Distribution Analysis: (a) Token count routed to each expert accumulated across all steps in DSMoE-S-E16. (b) Token count accumulated for different steps of selected experts.}
    \vspace{-4mm}
    \label{fig:4}
\end{figure}

\subsection{Temporal Similarity of Expert Features}

Existing DiT caching methods typically treat each token as the atomic reuse unit. The underlying rationale is that, across adjacent timesteps, most token-level features remain highly similar, while only a small subset of tokens undergo relatively large changes.

Starting from Fig.~\ref{fig:2}(a), we futher compare the temporal distances of token-level features and expert-branch features between adjacent timesteps, as illustrated in Fig.~\ref{fig:3}(a). 
    The results show that Top-$K$ expert routing introduces discontinuous changes in expert-branch features when activation states switch, while token-level variations closely follow these branch-level transitions. 
    In contrast, branches that remain activated across adjacent timesteps generally maintain high feature similarity. 
    Moreover, different branches exhibit distinct evolution patterns in both timing and magnitude. 
    These results indicate that DiT-MoE still contains substantial temporal redundancy, but this redundancy is no longer uniformly distributed across tokens; instead, it is structured along expert branches.

We further analyze the evolution of expert activation states across timesteps. 
    As shown in Fig.~\ref{fig:3}(b), we measure the routing weight distance across layers during sampling to characterize expert activation changes. 
    The results show that routing weights remain largely stable throughout most of the denoising process, with only occasional variations in early stages or local intervals. 
    This indicates that despite dynamic routing, expert branches preserve strong temporal consistency across adjacent timesteps, supporting the feasibility of branch-level feature reuse.

Therefore, the natural reuse unit in DiT-MoE is the expert branch rather than the entire token. 
    Token-level caching forces branches with different evolution patterns to share the same reuse decision, causing a mismatch between caching granularity and the underlying MoE computation structure.

\subsection{Spatial Heterogeneity Across Experts}

Beyond branch-level continuity across timesteps, from a structural perspective, MoE commonly exhibits clear expert specialization. 
    As shown in Fig.~\ref{fig:4}(a), we count how frequently each token is selected by different experts across all MoE layers throughout the denoising process, and map these statistics back to spatial positions.
    As shown in Fig.~\ref{fig:4}(b), we further examine how these activation patterns evolve along the denoising trajectory.
The results show that different experts exhibit markedly different spatial response patterns. 
    Some experts display more localized and concentrated activation distributions, mainly responding to regions with richer semantic variation or more complex structures. 
    In contrast, other experts respond over broader and smoother regions, covering more background areas or low-frequency structures. Meanwhile, their preferences over denoising stages are also different: some experts are more concentrated in specific stages, whereas others participate in representation updating over a much wider temporal range.

These observations indicate that different experts carry different feature components. More importantly, such structural differences suggest that caching control should not implicitly assume that all experts share the same tolerance to reuse error.

\subsection{Discussion}

The above analysis reveals that introducing MoE into DiT fundamentally reorganizes the evolution of noisy features during diffusion. 
    Temporal consistency and variation are no longer uniformly distributed across tokens but are instead manifested through branch-level updates of different experts. Therefore, the limitation of existing caching methods in DiT-MoE is not due to insufficient caching sophistication, but rather the mismatch between the caching granularity and the intrinsic feature evolution structure of DiT-MoE.

This observation leads to three requirements. First, caching should move from the token level to the expert-branch level to align the reuse unit with the computation unit. Second, caching decisions should account for expert-specific behaviors instead of applying uniform reuse strategies. Third, branch-wise updates require maintaining consistency among related subpaths to prevent representation drift caused by partial updates.
    Based on these insights, the next section introduces a fine-grained expert caching framework for DiT-MoE, enabling cross-timestep feature reuse at a granularity aligned with MoE-driven feature evolution.
\begin{figure*}[t!]
    \centering
    \includegraphics[width=6.9in]{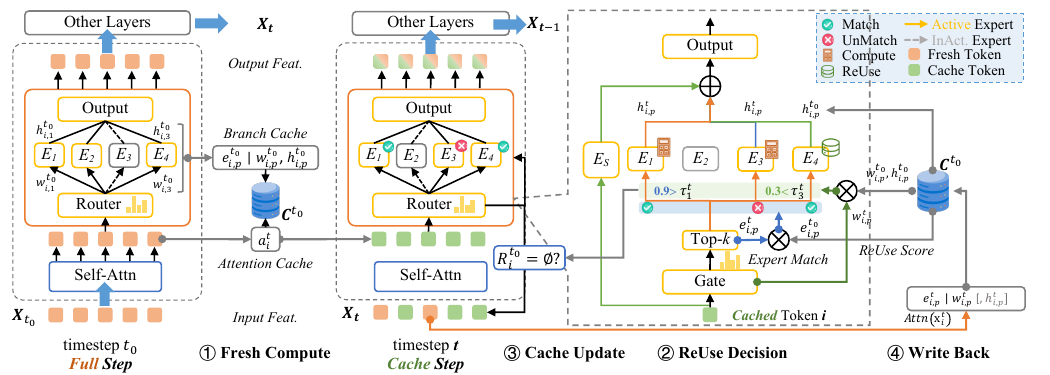}
    \vspace{-4mm}
    \caption{Overview of MoECa.}
    \label{fig:5}
    \vspace{-4mm}
\end{figure*}
\section{Methodology}

As analyzed in Section~3, existing caching methods in DiT-MoE suffer from a structural mismatch between the caching unit and the actual noisy feature evolution under expert decomposition. To address this issue, we propose \textbf{MoECa}, which performs caching and recomputation at the expert-branch level rather than the token level, aligning the reuse granularity with both MoE computation decomposition and branch-wise cross-timestep feature evolution.

\subsection{Expert Cache Architecture}

For the $i$-th token at the current denoising step $t$, let $t_0>t$ denote the most recent step whose branch states are currently stored in the cache for this token. We then define the current top-$k$ routing result and the cached reference state of token $i$ as:
\begin{equation}
\mathcal{B}_{i}^{t}=\{(e_{i,p}^{t}, w_{i,p}^{t})\}_{p=1}^{k}, 
\quad
\mathcal{C}_{i}^{t_0}=\{(e_{i,q}^{t_0}, w_{i,q}^{t_0}, h_{i,q}^{t_0})\}_{q=1}^{k},
\end{equation}
    For $\mathcal{B}_{i}^{t}$, $e_{i,p}^{t}$ denotes the index of the selected expert, $w_{i,p}^{t}$ denotes the corresponding routing weight, and $p$ indexes the routed branches within the token.
    For $\mathcal{C}_{i}^{t_0}$, the superscript $t_0$ explicitly marks the cached reference step rather than the current one, and $h_{i,q}^{t_0}$ is the output feature of the $q$-th cached expert branch.
    In addition to the MoE branch states, we also cache the token-wise attention output $a_i^{t_0}$ following the standard caching paradigm in DiT acceleration.

Similar to prior caching methods, MoECa alternates between \textit{full-computation steps} and \textit{cache steps}. 
    At each \textit{full-computation step}, the model executes the complete forward pass at step $t$ and overwrites the cache with the resulting $(e^{t},w^{t},h^{t},a^{t})$. 
    At each \textit{cache step}, the router is still executed normally to obtain the current routing result $\mathcal{B}_{i}^{t}$. 
    MoECa then compares the current branches with the cached branches in $\mathcal{C}_{i}^{t_0}$ and partitions them into a recomputation set $\mathcal{R}_{i}^{t}$ and a reuse set $\mathcal{B}_{i}^{t}\backslash \mathcal{R}_{i}^{t}$, as determined by the branch-matching and branch-scoring mechanism described in the next subsection.

A further issue is state consistency between the MoE path and the attention path.
    As shown in Section~3.1, changes in token features are highly aligned with changes in expert-branch activation. 
    If the expert features are partially updated while the attention feature remains stale, different sub-features of the same token would correspond to different timesteps, introducing additional representation drift.
    To avoid this inconsistency, MoECa coordinates the updates as follows:
        If no branch is selected for recomputation, both the MoE path and the attention path reuse their cached states. 
        Otherwise, MoECa recomputes only the branches in $\mathcal{R}_i^{t}$, reuses the remaining branches, and updates the attention feature at the same step.
   
Accordingly, the MoE output of token $i$ at step $t$ is written as
\begin{equation}
y_i^{t}
=
\sum\nolimits_{p\in \mathcal{R}_i^{t}} w_{i,p}^{t} h_{i,p}^{t}
+
\sum\nolimits_{p\in \mathcal{B}_{i}^{t}\backslash \mathcal{R}_{i}^{t}} w_{i,p}^{t} h_{i,q(p)}^{t_0},
\end{equation}
    where $q(p)$ denotes the cached branch matched to the current branch $p$. In other words, recomputed branches use the newly computed outputs at the current step $t$, while reusable branches directly read the matched historical features from the cached reference step.

\subsection{Expert-Level ReUse Decision}

At each \textit{cache} step, MoECa determines which routed branches require recomputation and which can reuse cached states. 
    To this end, MoECa first performs expert-branch matching and then assigns a branch-wise recomputation score.

For each current branch $p$, if its expert identity $e_{i,p}^{t}$ can be found in the cached state $\mathcal{C}_{i}^{t_0}$, we regard it as \emph{matched} and record the matched cached position as $q(p)$. 
    Otherwise, it is treated as \emph{unmatched}. 
    This design is consistent with the analysis that expert selection usually remains relatively stable across adjacent timesteps, with routing weights changing smoothly. 

For \textit{unmatched} branches, no semantically corresponding cached state is available. 
    Therefore, reusing historical features is unreliable, and these branches are directly assigned to the recomputation set.

For \textit{matched} branches, MoECa estimates the recomputation priority of each branch based on three signals:
\textbf{(1) Routing weight.}
    The routing weight $w_{i,p}^{t}$ indicates the current contribution of branch $p$ to token representation. 
    A larger weight implies higher importance and requires more conservative reuse decisions.
\textbf{(2) Temporal drift.}
    The temporal drift $\Delta_{i,p}^{t}$ measures the change in routing weight between the current branch and its cached counterpart. 
    Although the expert identity remains unchanged, a large drift indicates significant variation in activation strength and a higher risk of feature reuse.
\textbf{(3) Historical contribution.}
    We define $o_{i,q(p)}^{t_0}$ as the relative contribution of the matched cached branch in the reference state. 
    A branch with a larger historical contribution has a greater impact on the representation, and an incorrect reuse decision may therefore introduce larger errors.
Combining these signals, we define the \textbf{branch-level recomputation score} as $s_{i,p}^{t}$: 
\begin{equation}
\begin{gathered}
    \Delta_{i,p}^{t}=\left|w_{i,p}^{t}-w_{i,q(p)}^{t_0}\right|,\\
    o_{i,q(p)}^{t_0} =
        {\|h_{i,q(p)}^{t_0}\|}/
        {\sum\nolimits_{j}\|h_{i,j}^{t_0}\|},\\
    s_{i,p}^{t} =
        w_{i,p}^{t}
        +\lambda_{1}\Delta_{i,p}^{t}
        +\lambda_{2}o_{i,q(p)}^{t_0},
\end{gathered}
\end{equation}
where $\lambda_{1}$ and $\lambda_{2}$ control the relative importance of the temporal-drift term and the historical-contribution term. 
    We use the branch-output norm for $o$, rather than another smoothed routing weight, because routing weights encode router preference before the expert transformation, whereas reuse error depends on the realized feature contribution after the FFN. 
    Thus, branches with similar routing weights may still have substantially different effects on MoE output.

Then, for a \textit{matched} branch, the recomputation decision is made by thresholding this score, $p\in\mathcal{R}_{i}^{t} \Leftrightarrow s_{i,p}^{t}>\tau_{i,p}^{t}.$
In the basic form, $\tau_{i,p}^{t}$ can be a constant threshold. In the next subsection, we further extend it to an expert-aware adaptive threshold. Through this mechanism, MoECa aligns the caching decision with the actual computation decomposition of MoE, allowing different expert branches within the same token to be recomputed or reused independently.

\subsection{Expert-Aware Adaptive Threshold}

If a single threshold is applied throughout inference, the reuse criterion remains uniform even after refining the caching unit to the expert-branch level, ignoring the expert heterogeneity discussed in Section~3.2. 
    To incorporate this heterogeneity into the caching decision, we use the spatial response entropy of each expert as a proxy for its sensitivity to feature variation.
    Intuitively, experts with responses concentrated in limited spatial regions exhibit lower-entropy distributions and are more likely to capture local details. 
    Such experts are more sensitive to feature changes and therefore require more conservative reuse.

To estimate this behavior online, we maintain a runtime spatial activation statistic for each expert. Let $C_{e,i}^{t}$ denote the response statistic of expert $e$ on token position $i$ at step $t$. We update it as
\begin{equation}
C_{e,i}^{t}
=
\beta C_{e,i}^{t_0}
+
(1-\beta)\,\mathbb{I}\!\left[e\in\mathcal{B}_{i}^{t}\right],
\qquad \beta\in[0,1),
\end{equation}
where $\mathbb{I}[\cdot]$ is the indicator function and $\beta$ controls the exponential moving average.
Based on $C_{e,i}^{t}$, we construct the spatial response distribution of expert $e$ at step $t$ and compute its entropy as:
\begin{equation}
P_{e}^{t}(i) = 
    {C_{e,i}^{t}}/{\sum\nolimits_{j} C_{e,j}^{t}}, 
    \quad
H_{e}^{t} =
    -\sum\nolimits_{i} P_{e}^{t}(i)\log P_{e}^{t}(i).
\end{equation}
To make experts within the same layer comparable, we normalize the entropy and obtain $\bar H_{e}^{t}\in[0,1]$.

The expert-aware threshold is then defined by mapping the normalized entropy to a per-expert threshold:
\begin{equation}
    \tau_{e}^{t} =
        \tau_{0}\bigl(1+\alpha(2\bar H_{e}^{t}-1)\bigr), \qquad \alpha\in[0,1),
\end{equation}
where $\tau_{0}$ is the global base threshold and $\alpha$ controls the strength of expert adaptation. 
    When $\bar H_{e}^{t}$ is small, the threshold becomes lower and the corresponding branches are more likely to be recomputed. In contrast, the  branches are more likely to be reused.

\section{Experiments}

\begin{table*}[t]
    \centering
    \small
    \setlength\tabcolsep{11pt} 
    \caption{\textbf{Quantitative comparison on class-to-image generation} on ImageNet with \text{DSMoE.}}
    \vspace{-4mm}

\begin{tabular}{l | c c | c c | c | c | c | c}
\toprule

\bf{Method}  & \bf{Latency(ms)$\downarrow$} & \makecell{\bf{Latency}\\\bf{Speedup}$\uparrow$} & \bf{FLOPs(T)$\downarrow$} & \makecell{\bf{FLOPs}\\\bf{Speedup}$\uparrow$}  & \bf{FID$\downarrow$} &  \makecell{\bf{Inception}\\\bf{Score}$\uparrow$}  & \bf{PSNR$\uparrow$} & \bf{SSIM$\uparrow$} \\

\midrule

\textbf{DSMoE-S-E16}    & 64.32 & 1.00$\times$ & 53.62 & 1.00$\times$ & 39.82 & 38.57  & -- & -- \\
\textbf{33\% steps}     & 30.62 & 2.10$\times$ & 17.81 & 3.01$\times$ & 51.28 & 37.41  & 28.087 & 0.6776 \\
\textbf{37\% steps}     & 31.37 & 2.05$\times$ & 19.93 & 2.69$\times$ & 48.96 & 37.58  & 28.215 & 0.6831 \\
\textbf{ToCa}           & 33.19 & 1.94$\times$ & 20.25 & 2.65$\times$ & 41.16 & \textbf{38.70}  & 29.122 & 0.7082 \\
\textbf{DuCa}           & 32.78 & 1.96$\times$ & 20.31 & 2.64$\times$ & 40.83 & 38.64  & 29.276 & \underline{0.7137} \\
\textbf{TeaCache}       & 33.85 & 1.90$\times$ & 24.04 & 2.23$\times$ & \underline{40.59} & 38.62  & \underline{29.330} & 0.7097 \\
\rowcolor{gray!20}
\textbf{MoECa}          & 29.61 & 2.17$\times$ & 19.27 & 2.78$\times$ & \textbf{40.39} & \underline{38.67}  & \textbf{29.418} & \textbf{0.7142} \\

\midrule

\textbf{DSMoE-S-E48}    & 182.77& 1.00$\times$ & 43.25 & 1.00$\times$ & 40.21 & 38.31  & -- & -- \\
\textbf{33\% steps}     & 79.46 & 2.30$\times$ & 14.37 & 3.01$\times$ & 53.47 & 36.98  & 28.094 & 0.6784 \\
\textbf{37\% steps}     & 81.25 & 2.25$\times$ & 16.08 & 2.69$\times$ & 49.85 & 37.36  & 28.231 & 0.6847 \\
\textbf{ToCa}           & 86.26 & 2.12$\times$ & 16.51 & 2.62$\times$ & 41.62 & 38.26  & 29.087 & 0.7071 \\
\textbf{DuCa}           & 84.88 & 2.15$\times$ & 16.32 & 2.65$\times$ & 41.34 & \underline{38.28}  & \underline{29.314} & \underline{0.7136} \\
\textbf{TeaCache}       & 89.15 & 2.05$\times$ & 18.97 & 2.28$\times$ & \underline{40.89} & \underline{38.28}  & 29.286 & 0.7104 \\
\rowcolor{gray!20}
\textbf{MoECa}          & 79.17 & 2.31$\times$ & 15.38 & 2.81$\times$ & \textbf{40.58} & \textbf{38.35}  & \textbf{29.447} & \textbf{0.7145} \\

\midrule

\textbf{DSMoE-L-E16}    & 141.36 & 1.00$\times$ & 729.38 & 1.00$\times$ & 9.81 & 115.26    & -- & -- \\
\textbf{33\% steps}     & 70.67 & 2.00$\times$ & 242.31 & 3.01$\times$ & 16.37 & 113.84    & 28.784 & 0.7227 \\
\textbf{37\% steps}     & 71.15 & 1.99$\times$ & 271.14 & 2.69$\times$ & 14.26 & 114.12    & 28.961 & 0.7273 \\
\textbf{ToCa}           & 72.84 & 1.94$\times$ & 277.71 & 2.63$\times$ & 10.72 & \textbf{115.21}    & 29.861 & 0.7544 \\
\textbf{DuCa}           & 71.92 & 1.97$\times$ & 278.39 & 2.62$\times$ & 10.58 & 115.17    & 29.963 & 0.7656 \\
\textbf{TeaCache}       & 72.49 & 1.95$\times$ & 327.08 & 2.23$\times$ & \underline{10.47} & 115.14    & \underline{30.174} & \underline{0.7762} \\
\rowcolor{gray!20}
\textbf{MoECa}          & 68.71 & 2.06$\times$ & 263.39 & 2.77$\times$ & \textbf{10.21} & \underline{115.19}    & \textbf{30.227} & \textbf{0.7819} \\

\midrule

\textbf{DSMoE-L-E48}    & 352.32 & 1.00$\times$ & 617.83 & 1.00$\times$ & 9.20 & 118.42    & -- & -- \\
\textbf{33\% steps}     & 166.93 & 2.11$\times$ & 205.26 & 3.01$\times$ & 15.91 & 116.87   & 28.792 & 0.7231 \\
\textbf{37\% steps}     & 173.32 & 2.03$\times$ & 229.67 & 2.69$\times$ & 13.68 & 117.46   & 28.947 & 0.7265 \\
\textbf{ToCa}           & 177.20 & 1.99$\times$ & 236.71 & 2.61$\times$ & 10.14 & 118.37   & 29.836 & 0.7518 \\
\textbf{DuCa}           & 173.84 & 2.03$\times$ & 234.02 & 2.64$\times$ & 10.01 & 118.42   & 29.978 & 0.7642 \\
\textbf{TeaCache}       & 184.46 & 1.91$\times$ & 272.17 & 2.27$\times$ & \underline{9.76} & \underline{118.47}    & \textbf{30.238} & \underline{0.7776} \\
\rowcolor{gray!20}
\textbf{MoECa}          & 165.01 & 2.14$\times$ & 218.31 & 2.83$\times$ & \textbf{9.54} & \textbf{118.53}    & \underline{30.229} & \textbf{0.7808} \\

\bottomrule
\end{tabular}
\label{table:dsmoe}
\footnotesize\vspace{-3mm}
\end{table*}

\begin{table}[t]
    \centering
    \small
    \setlength\tabcolsep{5pt}
    \caption{Quantitative results with \text{DiT-MoE}.}
    \vspace{-3mm}

\begin{tabular}{l | c c c | c | c}
\toprule
\bf Method & \bf Latency(ms) $\downarrow$ & \bf FLOPs(T) $\downarrow$ & \bf Speed $\uparrow$ & \bf FID $\downarrow$ & \bf IS $\uparrow$ \\

\midrule

\textbf{DiT-MoE}    & 103.71 & 163.98 & 1.00$\times$ & 28.83 & 54.14 \\
\textbf{33\% steps} & 46.38 & 54.47 & 3.01$\times$ & 39.82 & 50.38 \\
\textbf{37\% steps} & 51.83 & 60.95 & 2.69$\times$ & 37.26 & 51.17 \\
\textbf{ToCa}       & 55.45 & 84.52 & 1.94$\times$ & 31.42 & 53.48 \\
\textbf{DuCa}       & 54.21 & 82.63 & 1.98$\times$ & 31.28 & 53.55 \\
\textbf{TeaCache}   & 54.87 & 91.61 & 1.79$\times$ & 31.13 & 53.52 \\
\rowcolor{gray!20}
\textbf{MoECa}      & 50.17 & 76.98 & 2.13$\times$ & 30.76 & 53.72 \\

\bottomrule

\end{tabular}
\label{tab:dit_moe_small}
\vspace{-3mm}
\end{table}

\begin{table}[t]
\centering
\footnotesize
\setlength{\tabcolsep}{4pt}
\caption{T2I task on HiDream-I1 at 1024px resolution.}
\vspace{-3mm}
\label{tab:t2i_exp}
\begin{tabular}{l|ccc|ccc}
\toprule
\textbf{Method}     & \textbf{Latency (s)}$\downarrow$ & \textbf{TFLOPs}$\downarrow$ & \textbf{Speedup}$\uparrow$ & \textbf{PSNR}$\uparrow$ & \textbf{SSIM}$\uparrow$ & \textbf{LPIPS}$\downarrow$ \\
\midrule
HiDream-I1 & 14.291 & 1242.431 & 1.00$\times$ & -- & -- & -- \\
ToCa       &  6.303 &  451.372 & 2.75$\times$ & 29.814 & 0.7124 & 0.3187 \\
DuCa       &  6.180 &  443.713 & 2.80$\times$ & 29.657 & 0.7059 & 0.3251 \\
TeaCache   &  5.876 &  481.562 & 2.58$\times$ & 29.184 & 0.7193 & \textbf{0.2973} \\
\rowcolor{gray!20}
\textbf{MoECa} & \textbf{5.612} & \textbf{424.169} & \textbf{2.93$\times$} & \textbf{29.923} & \textbf{0.7268} & 0.3026 \\
\bottomrule
\end{tabular}
\end{table}

\begin{figure}[t!]
    \centering
    \includegraphics[width=3.3in]{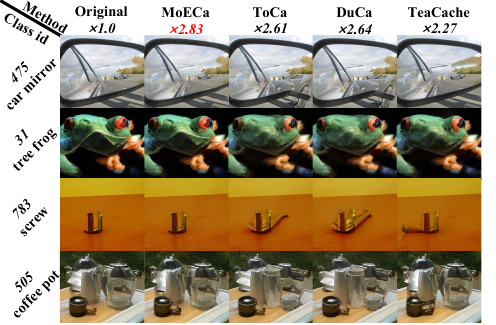}
    \caption{Qualitative results on DSMoE.}
    \vspace{-2mm}
    \label{fig:6}
\end{figure}

\subsection{Experimental Setup}

\noindent \textbf{Model Configuration.}
We evaluate MoECa on the class-conditional DSMoE~\cite{liu2025efficienttrainingdiffusionmixtureofexperts} and DiT-MoE~\cite{fei2024DiTMoE} families, and text-to-image (T2I) HiDream-I1~\cite{cai2025hidreami1}.
     DSMoE and DiT-MoE use 250 denoising steps at $256\times256$ resolution. 
     HiDream-I1 uses 50 sampling steps at $1024\times1024$ resolution on DrawBench200~\cite{saharia2022photorealistic}. 
     All latency measurements use an NVIDIA L40 GPU. We set the average forced refresh cycle to $N=4$ for MoECa and the token-level baselines. We select $\lambda_1$ and $\lambda_2$ through a small search on DSMoE-S-E48 and keep the selected configuration fixed for the remaining experiments.

\vspace{2pt}
\noindent \textbf{Evaluation Metrics.}
For class-conditional generation, we uniformly sample all 1,000 ImageNet categories and generate 50,000 images. We report FID and Inception Score (IS), together with PSNR and SSIM against dense outputs for DSMoE. For text-to-image generation, we use all DrawBench200 prompts and report latency, TFLOPs, PSNR, SSIM, and LPIPS. PSNR, SSIM, and LPIPS are computed between accelerated and dense outputs generated with matched prompts and random seeds.

\vspace{2pt} 
\noindent \textbf{Compared Methods.}
We compare with the dense model, reduced-step sampling, and three representative training-free caching methods: ToCa, DuCa, and TeaCache. ToCa and DuCa provide the closest token-level comparison, while TeaCache is a widely used step-level baseline. Quantization, pruning, and distillation alter weights, token sequences, or require training and are therefore complementary rather than directly matched feature-reuse baselines.

\begin{table}[t]
\centering
\caption{Ablation of adaptive control and sync updates.}
\vspace{-2mm}
\setlength\tabcolsep{6pt}
\small
\begin{tabular}{l | c c c c}
\toprule
\bf Method & \bf FID $\downarrow$ & \bf IS $\uparrow$ & \bf PSNR $\uparrow$ & \bf SSIM $\uparrow$ \\
\midrule
Fixed Threshold                  & 40.87 & 38.19 & 29.112 & 0.7082 \\
\textbf{Adaptive Threshold} ($\alpha$=0.4) & 40.72 & 38.27 & 29.376 & 0.7118 \\
\textbf{Adaptive Threshold} ($\alpha$=0.8) & 40.65 & 38.30 & 29.285 & 0.7131 \\
\textbf{Adaptive Threshold} ($\alpha$=0.6) & 40.58 & 38.35 & 29.447 & 0.7145 \\
\underline{Without Sync. Update} & 42.35 & -- & -- & 0.6918 \\
\bottomrule
\end{tabular}
\label{tab:ablation_threshold}
\vspace{-4mm}
\end{table}

\begin{figure}[b]
    \vspace{-3mm}
    \centering
    \includegraphics[width=3.3in]{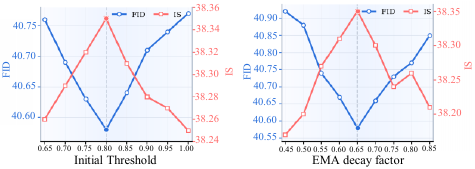}
    \vspace{-3mm}
    \caption{Parameter Ablation Experiment.}
    \label{fig:7}
\end{figure}

\subsection{Experiment Results}
\label{sec:exp_result}
\textbf{Main Results.} Tab.~\ref{table:dsmoe} and~\ref{tab:dit_moe_small} show a favorable speed--quality balance across five configurations. Among caching baselines, MoECa has the lowest latency and FLOPs and the best FID in each setting, although some pairwise FID gaps are small and should not be interpreted as statistical significance. On DSMoE-L-E48, MoECa reduces latency to 165.01 ms (2.14$\times$) and computation to 218.31 TFLOPs (2.83$\times$), with FID 9.54, achiving the most significant speedup of 2.83 $\times$.

\noindent \textbf{Comparison with Reduced Sampling-Step Methods.} 
    Although reducing sampling steps achieves higher theoretical speedups, aggressive step reduction leads to noticeable quality degradation. 
    For example, on DSMoE-L-E48, 33\% steps achieve a 3.01$\times$ speedup but increase FID to 15.91, while MoECa maintains an FID of 9.54 with a comparable 2.83$\times$ speedup. 
    Similar trends are observed across other settings, indicating that coarse-grained step reduction struggles to preserve intermediate denoising dynamics. 
    In contrast, MoECa selectively reuses expert branches, preserving critical representations while maintaining high acceleration.

\noindent \textbf{Comparison with Caching Methods.} 
    Compared with ToCa, DuCa, and TeaCache, MoECa reduces latency by 4.5\%--11.4\% over the strongest caching baseline and achieves the lowest FLOPs, with its generation quality still comparable to original generation.
    Despite that some FID differences are not statistically significant, the consistent efficiency improvement of MoECa under comparable quality demonstrates the advantage of branch-level caching.

\noindent \textbf{Trends Across Model Scales.} The E48 configurations yield slightly larger FLOPs speedups than E16 (2.81$\times$ versus 2.78$\times$ on DSMoE-S and 2.83$\times$ versus 2.77$\times$ on DSMoE-L). We attribute this trend to E48's finer decomposition into more, smaller experts, which offers more selectively reusable units. By comparison, DiT-MoE uses fewer, larger expert branches and attains a lower 2.13$\times$ FLOPs speedup. This cross-family gap is also affected by the different fractions of attention, shared-path, and routed-expert computation that branch caching can address, rather than expert count alone.

\noindent \textbf{Text-to-Image Generalization.} Tab.~\ref{tab:t2i_exp} extends the evaluation beyond ImageNet to general text-to-image task. 
    MoECa obtains the lowest latency and computation among the accelerated methods, providing 2.55$\times$ measured latency and 2.93$\times$ FLOPs speedups. It also obtains the best PSNR and SSIM. These results show that branch-level reuse remains effective with text conditioning and at substantially higher spatial resolution.

\noindent \textbf{Qualitative Results.} Figure~\ref{fig:6} presents a visual comparison of different methods. It can be observed that in high-frequency regions—such as the rearview mirror scene, frog textures, screw details, and the edges of coffee pot—the results generated by MoECa are closer to those of the original model. In contrast, ToCa, DuCa, and TeaCache exhibit more noticeable structural blurring, edge stretching, or loss of local details in some examples. This further verifies that MoECa is able to preserve local details and overall structural consistency well even under high-acceleration settings.

\begin{table}[t]
\centering
\caption{Ablation of branch-scoring signals.}
\label{tab:branch_scoring_ablation}
\vspace{-2mm}
\setlength\tabcolsep{10pt}
\small
\begin{tabular}{l|cccc}
\toprule
\textbf{Score} & \textbf{FID}$\downarrow$ & \textbf{IS}$\uparrow$ & \textbf{PSNR}$\uparrow$ & \textbf{SSIM}$\uparrow$ \\
\midrule
$w$ only & 41.74 & 37.63 & 28.801 & 0.7032 \\
$\Delta$ only & 41.36 & 37.79 & 28.948 & 0.7058 \\
$o$ only & 41.58 & 37.66 & 28.873 & 0.7046 \\
$w+\Delta$ & \underline{40.67} & \underline{38.24} & \textbf{29.462} & \underline{0.7136} \\
$w+o$ & 41.01 & 37.96 & 29.158 & 0.7089 \\
$\Delta+o$ & 40.84 & 38.11 & 29.287 & 0.7115 \\
\rowcolor{gray!20}
\textbf{All} & \textbf{40.58} & \textbf{38.35} & \underline{29.447} & \textbf{0.7145} \\
\bottomrule
\end{tabular}
\vspace{-4mm}
\end{table}

\begin{figure}[b]
    \centering
    \vspace{-3mm}
    \includegraphics[width=3in]{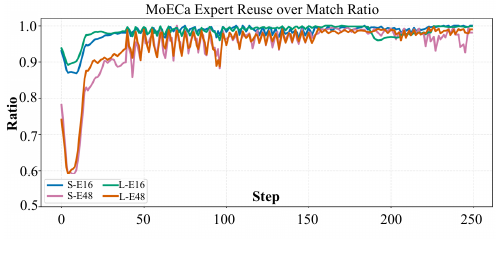}
    \vspace{-7mm}
    \caption{Reuse-over-match ratio (\%) across denoising steps.}
    \label{fig:8}
\end{figure}

\subsection{Ablation Study}
\label{sec:ablation}

\noindent \textbf{Ablation on the Adaptive Threshold.}
We compare fixed and adaptive thresholds to evaluate dynamic expert control. 
    The adaptive strategy ($\alpha=0.6$) achieves the best performance across all metrics, indicating that a uniform threshold cannot capture expert-specific variations. 
    Removing synchronized MoE/attention updates degrades performance, increasing FID from 40.58 to 42.35 and reducing SSIM from 0.7145 to 0.6918, demonstrating the importance of preventing temporally inconsistent states.
    
\noindent\textbf{Branch-scoring Signals.}
Tab.~\ref{tab:branch_scoring_ablation} isolates routing weight $w$, temporal drift $\Delta$, and historical contribution $o$. Single-signal variants are consistently weaker, while the complete score gives the best FID, IS, and SSIM, supporting their complementary roles.

\noindent \textbf{Impact of the Initial Base Threshold $\tau_0$.}
The base threshold $\tau_0$ controls the overall reuse aggressiveness. 
We evaluate $\tau_0$ from 0.65 to 1.00 and observe that generation quality first improves and then degrades as $\tau_0$ increases, with the best performance at $\tau_0=0.80$. 
A lower threshold causes excessive reuse and larger approximation errors, while a higher threshold triggers unnecessary recomputation and reduces caching efficiency. 
Thus, $\tau_0=0.80$ provides a better trade-off between reuse ratio and feature fidelity.

\noindent \textbf{Impact of EMA Decay Factor $\beta$.}
We employ an exponential moving average (EMA) to estimate expert-space response statistics, where $\beta$ controls the effective history length. 
We evaluate $\beta$ from 0.45 to 0.85 and achieve the best FID and IS performance at $\beta=0.65$. 
A smaller $\beta$ makes the statistics overly sensitive to local fluctuations, whereas a larger $\beta$ introduces delayed adaptation to routing changes. 
Therefore, a moderate decay factor provides a more stable and accurate estimation for adaptive cache control.

Figure~\ref{fig:8} shows that the reuse-over-match ratio remains high across all DSMoE variants and quickly stabilizes after early-stage fluctuations, indicating that most expert branches maintain consistent alignment across adjacent timesteps. 
    This observation validates the feasibility of branch-level reuse in DiT-MoE and supports the design of MoECa. 
    Compared with E16, E48 variants exhibit larger initial fluctuations, suggesting increased routing diversity and branch heterogeneity with more experts. 
    Nevertheless, the overall ratio remains high, indicating that richer expert decomposition provides greater opportunities for fine-grained selective reuse, which explains the larger speedup gains of MoECa under higher-expert configurations.

\subsection{Systems Overhead}

Memory consumption is measured on DSMoE-S-E48 after model loading and during one complete 250-step sampling run, using \texttt{torch.cuda.max\_memory\_allocated}.
    The latency breakdown is measured on DSMoE-S-E16 with the same setup, noting that component timings are profiled individually and are not necessarily additive.
Results in Tab.~\ref{tab:overhead} shows that, MoECa trades memory for finer-grained reuse. 
    On DSMoE-S-E48, its peak allocated memory is 3.303 GB, only 0.315 GB above DuCa despite storing branch states.
    On DSMoE-S-E16, the complete control logic takes only 4.8\% of the end-to-end time, remaining dominated by attention and expert computation. 
The current measurements cover single-GPU inference; expert-parallel load balance is left for future systems work.

\section{Related Works}

\subsection{Transformer and MoE in Diffusion Models}
Diffusion Transformers~\cite{peebles2023DiT} (DiTs) have emerged as a dominant paradigm in visual generation due to their remarkable ability to produce high-quality and diverse content.
    Recently, MoE~\cite{dai2024deepseekmoe} has emerged as an important new paradigm. By replacing part of the dense feed-forward layers with sparse expert modules and using a routing mechanism to dynamically activate only a small subset of experts for different tokens, it enables conditional computation. 
    In this way, DiT can reduce the number of activated parameters and the inference cost while maintaining or even improving model capacity and generation quality, thereby improving generation efficiency and model scalability. 
    For example, DiT-MoE~\cite{fei2024DiTMoE} scales sparse diffusion transformers to 16B parameters and reports expert preferences over spatial positions and denoising stages. We do not claim this specialization itself as new; MoECa instead studies and exploits its consequence for cross-timestep branch reuse.
    EC-DiT~\cite{sun2025ecdit} further designs an adaptive computation allocation mechanism based on expert-choice
    routing, scaling the model to 97B parameters and demonstrating strong scalability.
    DiffMoE~\cite{shi2025diffmoe} addresses the heterogeneity in the diffusion process by introducing a global token pool and a capacity predictor, making expert specialization and computation allocation more dynamic.
    These methods introduce or train new MoE architectures. In contrast, MoECa is a training-free inference method that retains a pretrained model's routing graph and selectively reuses intermediate branch features.

\subsection{Diffusion Model Acceleration}
To improve the generation efficiency of DiT models, many acceleration methods have been proposed, including quantization~\cite{chen2025qdit}, pruning~\cite{alvar2025divprune,chen2026toprovar}, and distillation~\cite{yin2024one}. Among them, cache-based methods have recently attracted increasing attention due to their training-free nature and strong generality. Existing methods can be broadly categorized into four groups: (1) Block-Level Caching: for example, DeepCache~\cite{ma2023deepcacheacceleratingdiffusionmodels} reduces computation by skipping and reusing the features of DiT blocks; (2) Layer-Level Caching: for example, $\Delta$-DiT~\cite{chen2024deltadittrainingfreeaccelerationmethod} and FORA~\cite{selvaraju2024fora} cache and reuse the features and residuals of different DiT layers; (3) Step Caching: for example, TeaCache~\cite{liu2024timestep} exploits the redundancy between adjacent diffusion timesteps and directly reuses historical results at timesteps with strong redundancy; and (4) Token-Level Caching: for example, ToCa~\cite{zou2024accelerating} and DuCa~\cite{zou2025rethinkingtokenwisefeaturecaching} further leverage the variation differences among tokens during the diffusion process, and only recompute a small number of critical tokens.
However, the granularity of these methods still remains at the block, layer, step, or token level, without being specifically designed for MoE architecture. 

\begin{table}[t]
\centering\scriptsize\setlength{\tabcolsep}{2.2pt}
\caption{Memory (Left) and Latency (Right) Overhead. }
\label{tab:overhead}
\vspace{-2mm}

\begin{minipage}[b]{0.54\linewidth}
\centering
\resizebox{\linewidth}{!}{
\begin{tabular}{lcc}
\hline
    \multirow{2}{*}{\textbf{Method}}
    & \textbf{Peak Mem.} & \textbf{Reserved} \\
    & \textbf{GB / $\Delta$} & \textbf{GB} \\
\hline
    Dense    & 0.780 / N/A         & 1.141 \\
    TeaCache & 0.964 / +23.6\%     & 1.159 \\
    ToCa     & 2.602 / +233.7\%    & 2.898 \\
    DuCa     & 2.648 / +239.7\%    & 2.988 \\
    MoECa    & 2.821 / +261.7\%    & 3.303 \\
\hline
\end{tabular}
}
\end{minipage}
\hfill
\begin{minipage}[t]{0.45\linewidth}
\centering
\resizebox{\linewidth}{!}{
\begin{tabular}{lcc}
\hline
    \textbf{Component} & \textbf{ms} & \textbf{Ratio} \\
\hline
    Full Model          & 71.772 & 100.0\% \\
    MoECa Overhead      & 3.502  & 4.8\% \\
    \midrule
    Scoring             & 1.634  & 2.3\% \\
    Router Execution    & 1.963  & 2.7\%\\
    Cache Update        & 0.616  & 0.8\% \\
    Entropy Calc.       & 1.235  & 1.7\% \\
\hline
\end{tabular}
}
\end{minipage}
\vspace{-4mm}
\end{table}
\section{Conclusion}

We present MoECa, a fine-grained caching framework for DiT-MoE that performs feature reuse at the expert-branch level instead of the whole-token level, enabling effective cross-timestep reuse through branch-level caching and expert-aware adaptive control. 
   Experiments on various DiT-MoE models demonstrate a favorable speed--quality trade-off. 
   Our current study focuses on single-GPU image generation.
   Extension to video and expert-parallel serving, where selective recomputation may interact with load balance, remains future work.


\balance
\bibliographystyle{ACM-Reference-Format}
\bibliography{
    _ref/ref
}

\end{document}